\documentclass[fleqn,moreauthors,10pt]{ds_report}
\newcommand{\keywordname}{Keywords}
\usepackage[english]{babel}

\graphicspath{{fig/}}

\JournalInfo{UL FRI Data Science }

\Archive{Reproducibility Study}

\PaperTitle{Reproducibility Study of: Large Language Model Bayesian Optimization}

\Authors{Adam Rychert\textsuperscript{1}, Gasper Spagnolo\textsuperscript{2} and Evgenii Posashkov\textsuperscript{3}}
\affiliation{\textsuperscript{1}\textit{ar77558@student.uni-lj.si}}
\affiliation{\textsuperscript{2}\textit{gs0104@student.uni-lj.si}}
\affiliation{\textsuperscript{3}\textit{ep84586@student.uni-lj.si}}


\Abstract{
In this reproduction study, we revisit the LLAMBO framework of Daxberger et~al.\ (2024), a prompting-based Bayesian optimization (BO) method that uses large language models as discriminative surrogates and acquisition optimizers via text-only interactions.
We replicate the core Bayesmark and HPOBench experiments under the original evaluation protocol, but replace GPT-3.5 with the open-weight Llama~3.1~70B model for all text-encoding components.

Our results broadly confirm the main claims of LLAMBO. Contextual warmstarting via textual problem and hyperparameter descriptions substantially improves early-regret behaviour and reduces variance across runs. LLAMBO’s discriminative surrogate is weaker than GP or SMAC as a pure single-task regressor, yet benefits from cross-task semantic priors induced by the language model. Ablations that remove textual context markedly degrade predictive accuracy and calibration, while the LLAMBO candidate sampler consistently generates higher-quality and more diverse proposals than TPE or random sampling. Experiments with smaller backbones (Gemma~27B, Llama~3.1~8B) yield unstable or invalid predictions, suggesting insufficient capacity for reliable surrogate behaviour.

Overall, our study shows that the LLAMBO architecture is robust to changing the language-model backbone and remains effective when instantiated with Llama~3.1~70B. All code and configurations are available at \url{https://github.com/spagnoloG/llambo-reproducibility}.
}

\begin{document}

\flushbottom

\maketitle

\maketitle
\thispagestyle{empty}


\section*{Introduction}
Black-box function optimization appears in many real-world problems such as robotics, experimental design, drug discovery, interface design, and, in machine learning, \emph{hyperparameter tuning}. Bayesian Optimization (BO) is a standard approach in this setting: it builds a surrogate model from past evaluations and uses an acquisition strategy to propose promising candidates, requiring only function evaluations but no gradients or closed-form objective. However, BO is often used in regimes where data are extremely sparse, so performance depends critically on search efficiency, the quality of the surrogate under few observations, and how well prior knowledge can be incorporated across tasks.

These few-shot challenges align naturally with the strengths of large language models (LLMs). Modern LLMs are trained on massive text corpora and show strong abilities in few-shot reasoning, pattern recognition, and contextual understanding. LLAMBO explores whether these capabilities can be used to support or even replace parts of the BO pipeline by expressing the entire loop in natural language. In the original work, an LLM is prompted with dataset and model metadata, past evaluations, and the current history, and is then asked to warmstart the search, act as a discriminative surrogate, and propose new hyperparameter candidates without fitting a conventional probabilistic model.

This paper presents a reproducibility study of LLAMBO in an open-model setting. We reconstruct the prompting pipeline, apply it to hyperparameter tuning tasks from Bayesmark and HPOBench, and compare performance against classical BO baselines such as GP-, SMAC-, and TPE-based optimizers. In contrast to the original study, which relies on GPT-3.5, we replace the backbone with open-weight Llama~3.1~70B and briefly probe smaller models as well. This allows us to assess both how well the original findings carry over and how sensitive LLAMBO is to the choice and capacity of the underlying language model.

\section*{Scope of Reproducibility}    
In this study, we aim to reproduce the main findings reported in the LLAMBO paper \cite{Liu2024LLMBO}, which tests whether large language models can replace key components of a Bayesian optimization loop when everything is expressed in natural language. LLAMBO claims that an LLM, prompted with dataset and model metadata plus past evaluations, can warmstart the search, propose new hyperparameter candidates, and estimate their performance without fitting a conventional surrogate model.

Our reproduction focuses on the following aspects:
\begin{itemize}
    \item \textbf{Warmstarting efficiency:} verify that contextual, text-based warmstarts achieve lower early regret and reduced variance compared to standard space-filling designs (Random, Sobol, Latin Hypercube).
    \item \textbf{Surrogate behaviour and calibration:} compare \\
    LLAMBO’s discriminative surrogate to GP- and RF-based baselines (GP, SMAC) in terms of predictive accuracy (NRMSE, $R^2$, regret) and uncertainty quality (LPD, coverage, sharpness).
    \item \textbf{Role of textual context:} assess the contribution of problem descriptions and hyperparameter-name embeddings via ablations that remove these signals, and measure the impact on prediction error and calibration.
    \item \textbf{Candidate generation quality:} evaluate the LLAMBO candidate sampler against TPE (independent and multivariate) and random sampling using regret-based metrics and diversity measures (generalized variance, log-likelihood).
\end{itemize}

We build directly on the original LLAMBO codebase, replacing all OpenAI API calls with local ollama \cite{ollama} invocations so that the full pipeline can be run with open-weight models. The implementation is organised primarily through nested bash scripts, with prompts and configuration spread across multiple files, and the repository does not provide utilities for aggregating results or plotting figures. To support this study, we implemented our own evaluation and plotting scripts for Bayesmark and HPOBench, standardised the LLM outputs into a common JSON format, and reimplemented all baselines (GP, SMAC, TPE, Random) using established libraries in a unified evaluation pipeline. In addition to the original GPT-3.5 setting, we test LLAMBO with Llama~3.1~70B as the main backbone and briefly probe smaller open models, allowing us to evaluate both the original claims and the method’s robustness to changes in the underlying language model.


\section*{Methodology}
\begin{figure}[h]
\centering
\includegraphics[width=1\columnwidth]{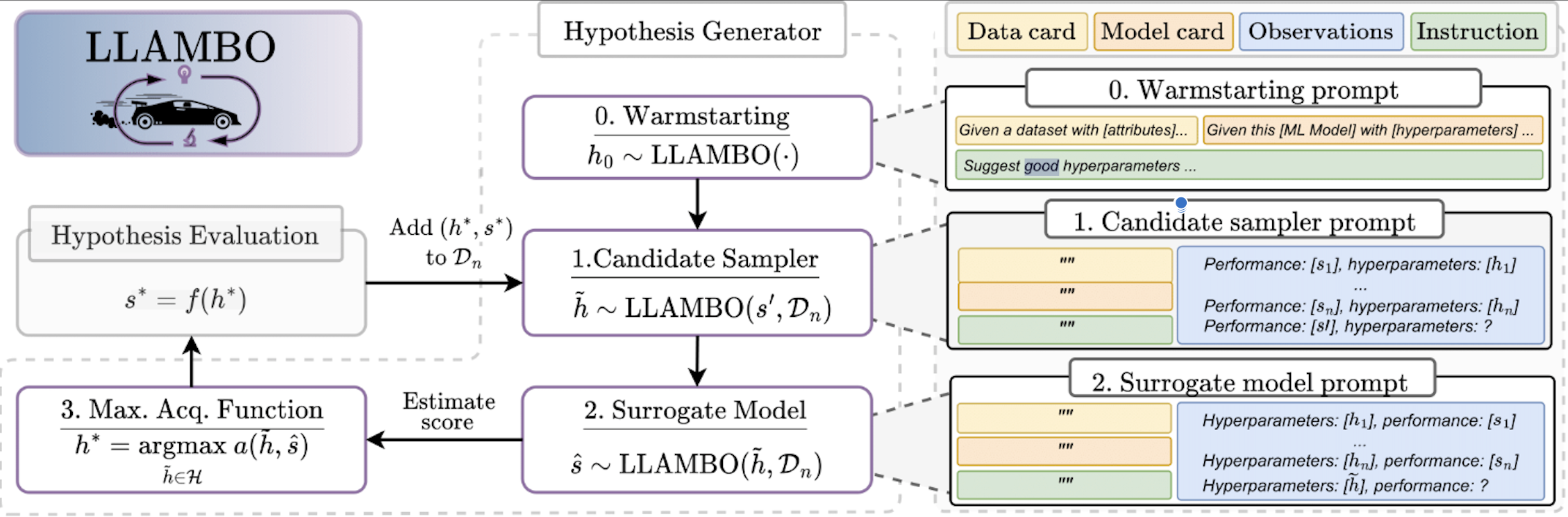}
\caption{\textbf{Overview of the LLAMBO prompting pipeline.}
A dataset description (Data Card), model specification (Model Card), prior observations, and task instructions are combined into structured prompts that guide the LLM through (0)~zero-shot warmstarting, (1)~candidate generation, and (2)~surrogate-based performance estimation. Each proposed hyperparameter configuration is evaluated by the objective function, added to the data history, and reintroduced into subsequent prompts, enabling iterative refinement of the search strategy purely through natural-language interactions.}
\label{fig:llambo_overview}
\end{figure}

\subsection*{LLAMBO Architecture}

LLAMBO replaces the usual components of Bayesian Optimization—surrogate models and acquisition functions—with a large language model that operates entirely through structured prompts. Instead of fitting a GP or a Random Forest and optimizing an acquisition function, the system repeatedly asks the LLM to propose hyperparameters, reason about previously observed trials, and estimate the performance of new candidates. Figure~\ref{fig:llambo_overview} illustrates the full loop.

At each step, LLAMBO interacts with Bayesmark in a closed loop: the LLM proposes a configuration, Bayesmark evaluates it, and the result is inserted into the next prompt. The prompting pipeline is built around three stages:

\begin{itemize}
    \item \textbf{Zero-shot warmstarting}
    \item \textbf{Candidate generation}
    \item \textbf{Surrogate-style performance estimation}
\end{itemize}

All prompts share two core pieces of information:

\begin{itemize}
    \item \textbf{Data Card} — dataset metadata (feature dimensionality, feature types, task type, etc.)
    \item \textbf{Model Card} — the hyperparameter search space for the current model class
\end{itemize}

These cards ensure the LLM always has access to the problem context. As new trials are evaluated, their scores and hyperparameters are appended to the prompt. The LLM is instructed to return results in a fixed JSON-like format so that outputs can be parsed directly into Bayesmark.

\subsubsection*{Zero-Shot Warmstarting}
In the first stage, LLAMBO asks the LLM to provide an initial hyperparameter configuration without seeing any previous evaluations \cite{Mallik2024Warmstarting}. The prompt contains only the Data Card, the Model Card, and a short instruction describing the goal. Based solely on this information, the LLM proposes a starting point. This acts as a drop-in replacement for the random, Sobol, or Latin Hypercube initializations used in standard BO, but relies entirely on the LLM’s interpretation of the dataset and model description rather than a fitted surrogate.

\subsubsection*{Candidate Generation}
After the first evaluation, LLAMBO enters the iterative candidate-generation loop. At each iteration, the LLM is prompted with the dataset and model information together with all previously observed (hyperparameter, performance) pairs. These observations are supplied in a simple text format. 
The LLM is then asked to suggest new configurations that might improve performance. Unlike classical BO, no explicit acquisition function is provided—the LLM decides which regions of the search space to explore or exploit based on the textual history.

\subsubsection*{Surrogate-Based Performance Estimation}
In the final stage, the LLM is asked to estimate the performance of a candidate before Bayesmark evaluates it. The prompt again contains the dataset metadata, model details, and the full evaluation history. The LLM outputs a numerical score representing its expectation of how the candidate will perform. In a conventional BO loop, this role is played by a trained surrogate (e.g., GP or Random Forest). Here, the LLM provides these predictions directly through pattern recognition in the textual description of the task and the observed history. LLAMBO then uses these estimated scores to rank candidates and select which one to evaluate next.

\subsection*{Datasets}
Our experiments follow the original LLAMBO setup and use the five tabular datasets included in the Bayesmark hyperparameter optimization benchmark. These datasets form the basis of the Data Card component shown in Figure~\ref{fig:llambo_overview}. Each dataset is provided in a standardized format that includes the feature matrix $X$, labels $y$, train–test splits, and metadata describing feature types, dimensionality, and task specification. This structure allows the dataset information to be inserted directly into the Data Card for every prompt, ensuring that the LLM always has access to the relevant problem context.

The five datasets cover a range of supervised learning problems. The Breast Cancer dataset is a binary classification task with 569 samples and 30 continuous features, making it suitable for evaluating LLAMBO’s warmstarting behaviour. The Diabetes dataset provides a regression problem with 442 instances and 10 clinical predictors, which is challenging for early surrogate estimation. The Digits dataset contains 1{,}797 samples of handwritten digits represented by 64 pixel-intensity features and tests LLAMBO’s ability to navigate a higher-dimensional multiclass setting. The Iris dataset is small and low-dimensional (150 samples, four botanical measurements), providing a scenario where the model must rely more heavily on prior knowledge. Finally, the Wine dataset contains 178 instances and 13 chemical attributes and serves as a moderately sized multiclass benchmark. Combined with the five Bayesmark model classes (Random Forest, AdaBoost, SVM, Logistic Regression, and a simple neural network), these datasets define the 25 optimization tasks reproduced in our study.

In the LLAMBO pipeline, each dataset enters the optimization loop through the Data Card and conditions all three components of the hypothesis generator. During zero-shot warmstarting, the Data Card is paired with the Model Card and instructions that guide the LLM to propose an initial hyperparameter configuration based on dataset characteristics such as dimensionality, class structure, and feature types. As the optimization progresses, the same dataset description is used in the candidate-generation prompt, where the LLM receives a history of evaluated configurations (``performance: $s_i$, hyperparameters: $h_i$'') and reasons about which regions of the search space may be promising. In the surrogate-estimation stage, the LLM again leverages the dataset metadata to predict how new candidates might perform, interpreting interactions between hyperparameters and dataset structure. For consistent evaluation across all methods, LLAMBO also uses Bayesmark’s precomputed performance statistics to compute regret in a standardized way.

\subsection*{Baseline Optimizers}

To evaluate LLAMBO in a full end-to-end hyperparameter optimization (HPO) setting, 
the original paper benchmarks the method against four widely used and methodologically 
diverse baseline optimizers. These baselines represent the dominant approaches in 
modern surrogate-based Bayesian Optimization, including classical Gaussian Processes(GPs), 
neural-augmented GPs, density-estimation methods, and random-forest surrogates. 
All baselines are run under identical conditions to ensure a fair comparison.

\paragraph{GP-DKL (Deep Kernel Learning Gaussian Process).}
GP-DKL is an advanced Gaussian Process model implemented in BoTorch. 
It combines a deep neural network with a GP kernel: the network learns 
a nonlinear feature embedding, and the GP operates on this learned space \cite{Wistuba2021FewShotDKS}. 
This hybrid surrogate retains calibrated uncertainty while offering 
greater flexibility in medium- and high-dimensional search spaces. 
Because of its strong modeling capacity, GP-DKL is considered one of the most 
powerful GP-based baselines in modern Bayesian Optimization pipelines.

\paragraph{SKOpt (Gaussian Process from Scikit-Optimize).}
It provides a classical implementation of Gaussian Process Bayesian Optimization \cite{Pedregosa2011ScikitLearn}. 
It uses stationary kernels (typically Matern 5/2) and standard acquisition 
functions such as Expected Improvement or Lower Confidence Bound. 
This baseline is known for its stability, interpretability, and reliable 
performance on low-dimensional and smooth objective landscapes, making it 
a canonical reference point in HPO studies.

\paragraph{Optuna (Tree-structured Parzen Estimator).}
Optuna’s default optimizer implements the Tree-structured Parzen Estimator (TPE), 
a non-parametric density-based approach to Bayesian Optimization. 
Rather than modeling the objective function directly, TPE models two conditional densities: one over high-performing configurations and one over low-performing ones. 
New candidates are drawn from regions where the ratio of these densities is favorable. 
This method scales well, handles categorical and conditional search spaces naturally, 
and is one of the most widely used HPO algorithms in practical machine learning systems \cite{Akiba2019Optuna}.

\paragraph{SMAC3 (Random-Forest Surrogate).}
SMAC3 uses a Random Forest regression model as its surrogate, with uncertainty 
estimated from tree-wise variance. This makes the method robust to non-smooth, 
noisy, and heterogeneous response surfaces. SMAC has a long history of strong 
performance in AutoML, and excels especially in hierarchical or irregular 
search spaces where GP-based methods may struggle \cite{Lindauer2022SMAC3}.

\paragraph{Evaluation Setup.}
All optimizers, including LLAMBO, are evaluated under the same conditions:
\begin{itemize}
    \item \textbf{5 randomly sampled initial points}
    \item \textbf{25 optimization trials} after initialization
    \item \textbf{5 independent runs} per task to reduce variance
\end{itemize}
The experiments cover a total of 30 tasks:
\begin{itemize}
    \item \textbf{25 Bayesmark tasks} (five datasets crossed with five model classes),
    \item \textbf{3 private datasets} not seen during LLM pretraining,
    \item \textbf{2 synthetic datasets} designed to probe behavior on 
    controlled objective landscapes.
\end{itemize}

This unified evaluation protocol ensures that differences in performance 
reflect the behavior of the optimization strategies themselves rather than 
differences in initialization or experimental configuration.

\section*{Results}
In this section, we present a comprehensive reproduction of the LLAMBO framework using the Llama 3.1 70B model as the backbone for all textual encodings. 
Our goal was to validate the core claims of the original paper regarding warmstarting efficiency, surrogate modeling behavior, the role of contextual embeddings,
and candidate point sampling. The reproduced results consistently align with the reported trends: contextual information embedded via the Llama 3.1 70B encoder provides a strong prior that substantially accelerates optimization, improves cross-task structure, and enables high-quality candidate generation. Across Figures, we observe the same characteristic signature of LLAMBO—effective warmstarts, meta-learned structure in surrogate predictions, clear degradation when textual information is removed, and state-of-the-art candidate quality—demonstrating that LLAMBO’s behavior is robust even under a different, larger language model backbone.

\subsection*{Baseline Optimizers}
Before showing the reproduced LLAMBO results, we made sure that our baseline implementation was completely in line with the setup described in \cite{Liu2024LLMBO}. We followed the original baseline protocol exactly to make sure that the methods were the same.  This verification step showed that our implementation works the same way as the baselines that were published and the results from a parallel reproduction.

After confirming equivalence, we also implemented a per-task min-max normalization scheme. Bayesmark's global scaling sets limits on all datasets, so regret curves never reach zero. In contrast, per-task normalization changes the performance of each task based on its observed minimum and maximum. This creates regret curves that stop at zero once the best configuration has been seen. This shows bigger differences between optimizers at the beginning while keeping the overall ranking the same. Importantly, we found that both normalization choices lead to the same qualitative ordering of the baselines. 

\begin{figure}[ht]
\centering
\includegraphics[width=0.9\columnwidth]{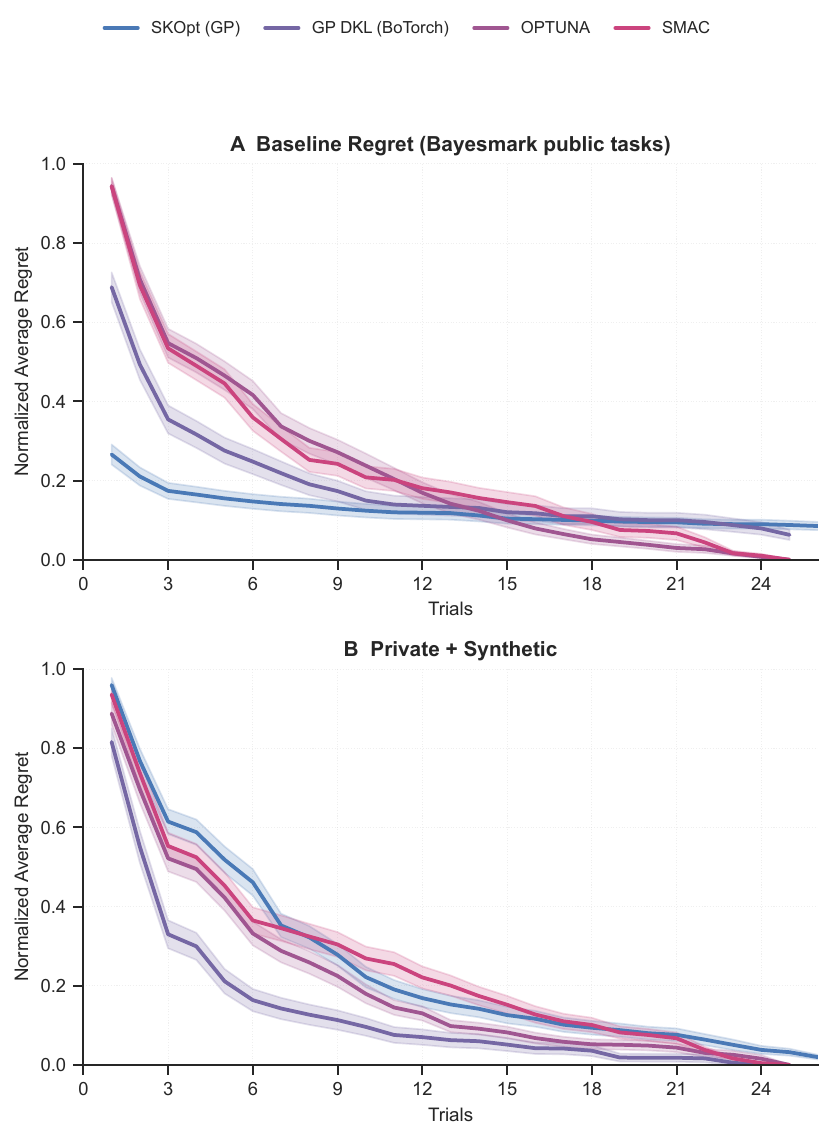}
\caption{\textbf{Baseline Regret on Bayesmark Public Tasks and  Private + Synthetic Tasks.}}
\label{fig:baseline_public}
\end{figure}

Figures~\ref{fig:baseline_public} offers an additional analysis to the global normalization curves. The original evaluation employs fixed Bayesmark-wide score bounds; however, our supplementary per-task min-max scaling uncovers more pronounced early-regret disparities and generates regret curves that plateau at zero upon reaching the task-specific optimum. This demonstrates that they are stable under alternative normalization schemes. It also highlights the importance of considering normalization choices when comparing Bayesian optimization methods.

\subsection*{Warmstarting strategies in Bayesian Optimization}

\begin{figure}[h]
\centering
\includegraphics[width=1\columnwidth]{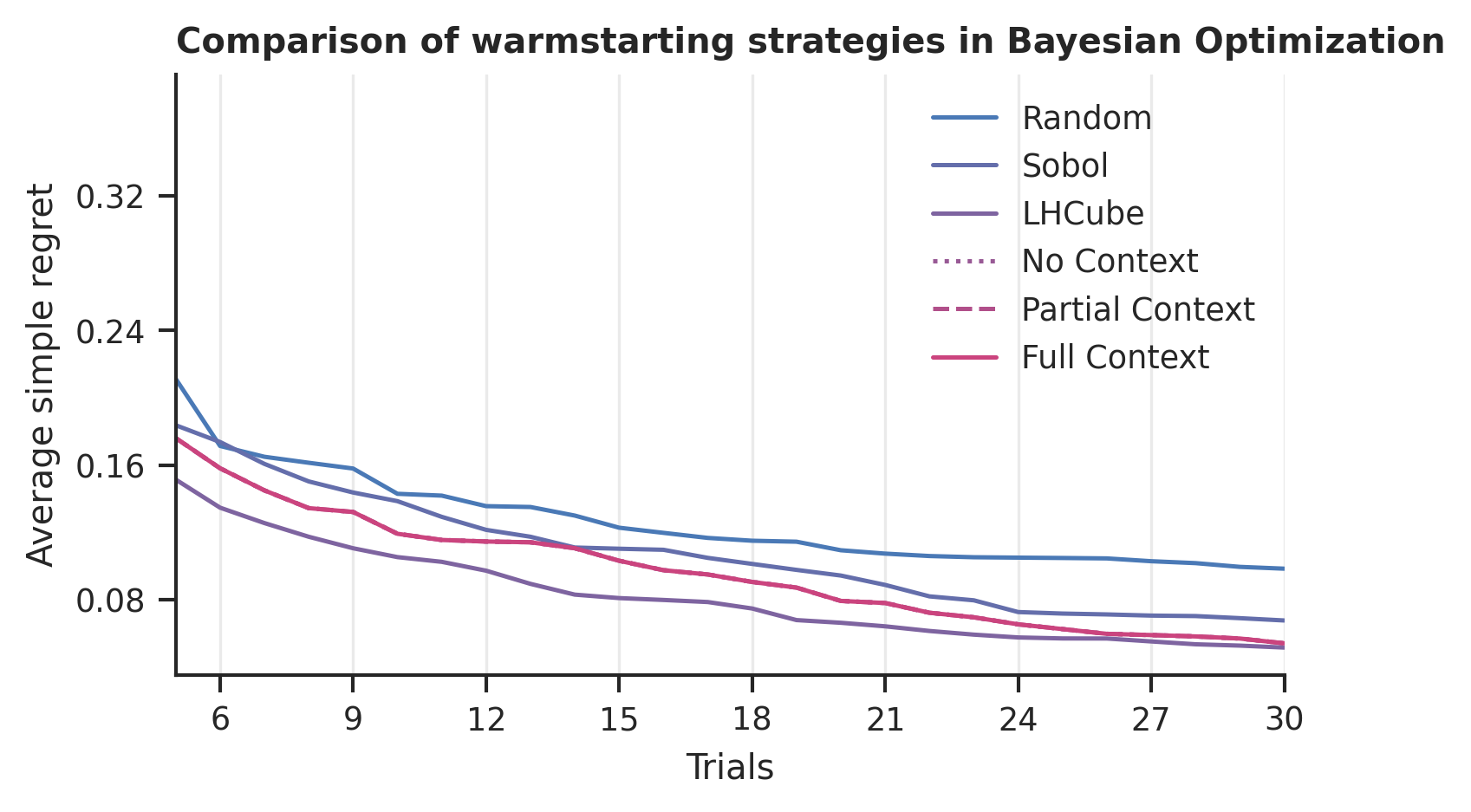}
\caption{\textbf{Warmstarting regret curves.} Average simple regret over trials for classical space-filling methods (Random, Sobol, LHCube) and contextual warmstarts (No Context, Partial Context, Full Context).}
\label{fig:warmstarting_regret}
\end{figure}

\begin{figure}[h]
\centering
\includegraphics[width=1\columnwidth]{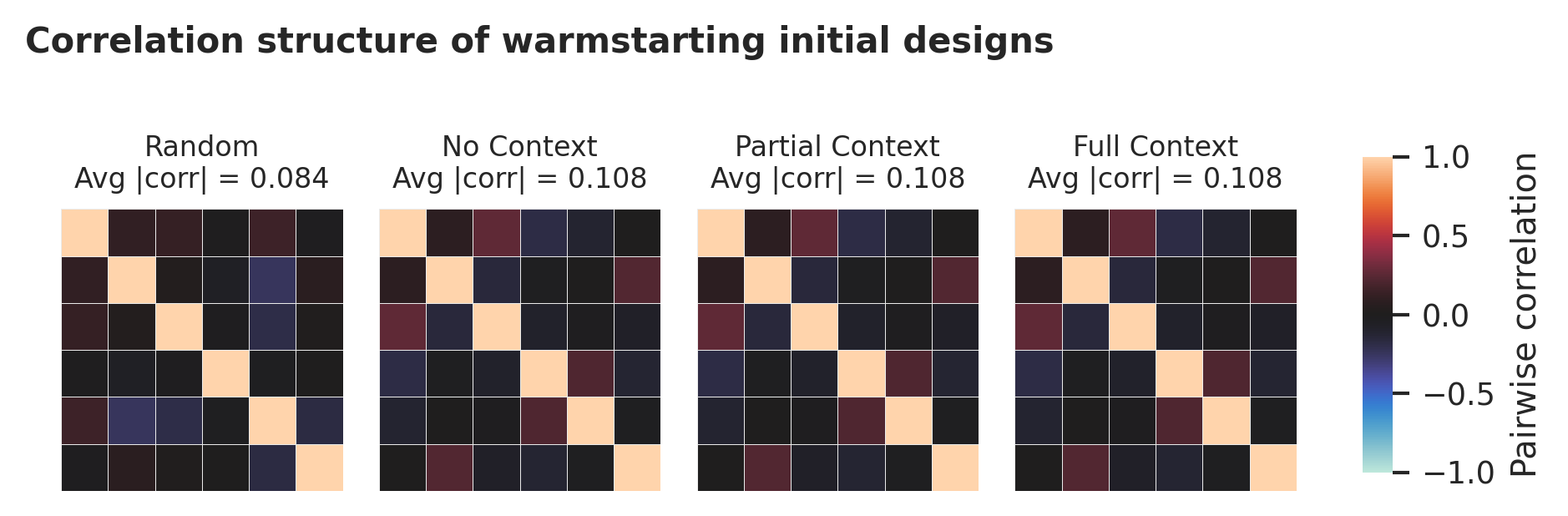}
\caption{\textbf{Correlation structure of initial designs.} Pairwise correlation matrices of normalized hyperparameters for different warmstarting strategies on a representative dataset/model.}
\label{fig:warmstarting_correlation}
\end{figure}

\begin{figure}[h]
\centering
\includegraphics[width=1\columnwidth]{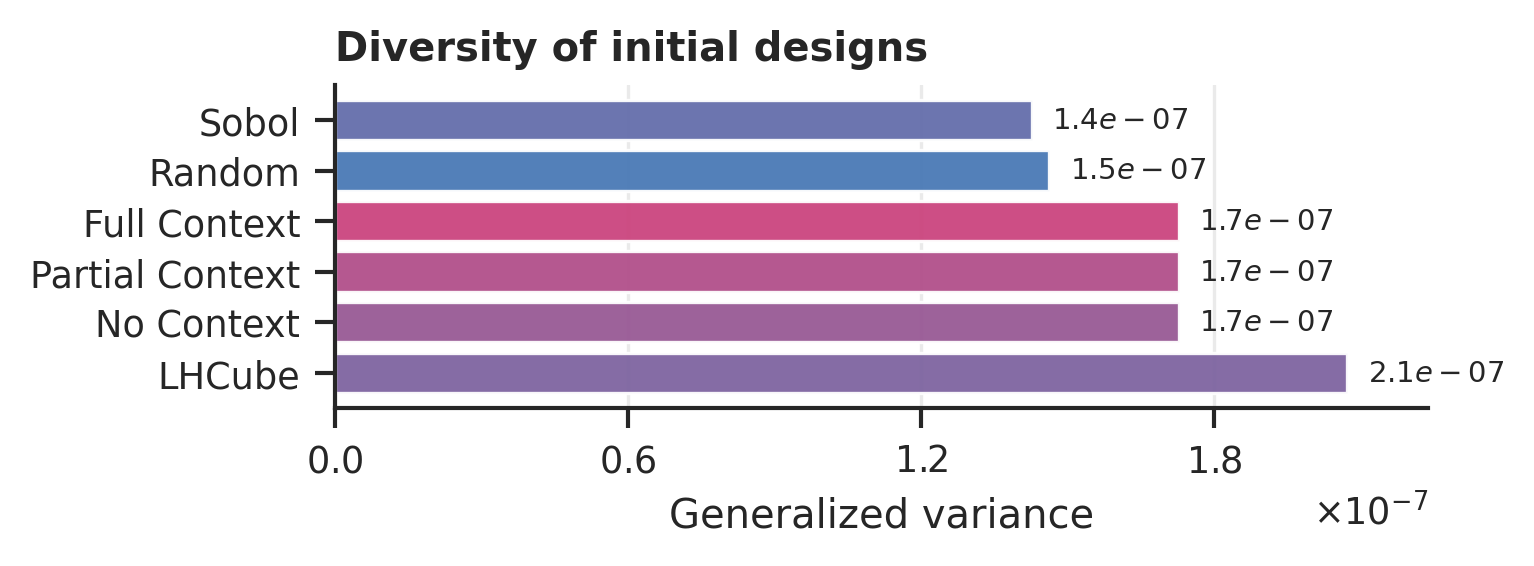}
\caption{\textbf{Diversity of initial designs.} Generalized variance of normalized hyperparameters for each warmstarting strategy (higher is more diverse).}
\label{fig:warmstarting_diversity}
\end{figure}

Figures~\ref{fig:warmstarting_regret}–\ref{fig:warmstarting_diversity} report the reproduced comparison of warmstarting strategies in Bayesian Optimization using the LLAMBO framework with a Llama~3.1~70B text encoder. We compare classical space-filling initialization schemes (Random, Sobol, Latin Hypercube) against contextual warmstarts that leverage problem descriptions and hyperparameter semantics (No Context, Partial Context, Full Context).

Figure~\ref{fig:warmstarting_regret} shows the average simple regret over optimization trials. The reproduced curves match the qualitative behavior of the original work: classical designs (Random, Sobol, LHCube) yield consistently higher regret, especially in the early stages, indicating weaker priors and slower convergence towards the task optimum. In contrast, contextual warmstarts exhibit uniformly lower regret throughout the optimization horizon, with the Full Context variant performing best. The shaded regions further indicate reduced variability across runs for contextual methods, confirming a more stable and reliable optimization trajectory.

To better understand how these warmstarts populate the search space, Figure~\ref{fig:warmstarting_correlation} visualizes the pairwise correlation structure of the initial designs for a representative task (\textit{breast}, RF). Random sampling yields the lowest average absolute correlation, consistent with nearly independent draws. In contrast, contextual warmstarts induce more structured correlation patterns between hyperparameters, reflecting task-specific priors encoded by the language model rather than purely independent sampling.

Figure~\ref{fig:warmstarting_diversity} complements this view by quantifying the overall diversity of initial designs via the generalized variance of the normalized hyperparameters. Latin Hypercube sampling achieves the highest diversity, as expected from a stratified design. However, contextual warmstarts attain diversity levels comparable to LHCube and substantially higher than Random and Sobol, while still encoding meaningful structure. This indicates that contextual initialization is not merely collapsing onto a narrow region of the search space, but instead produces diverse yet semantically informed starting points.

Overall, this reproduction confirms that warmstarting with contextual embeddings substantially improves sample efficiency and early-regret performance in Bayesian Optimization, while maintaining high diversity in the initial design and introducing task-aware structure into the explored hyperparameter space.

\subsection*{Discriminative Surrogate Models evaluation}
\begin{figure}[h]
\centering
\includegraphics[width=1\columnwidth]{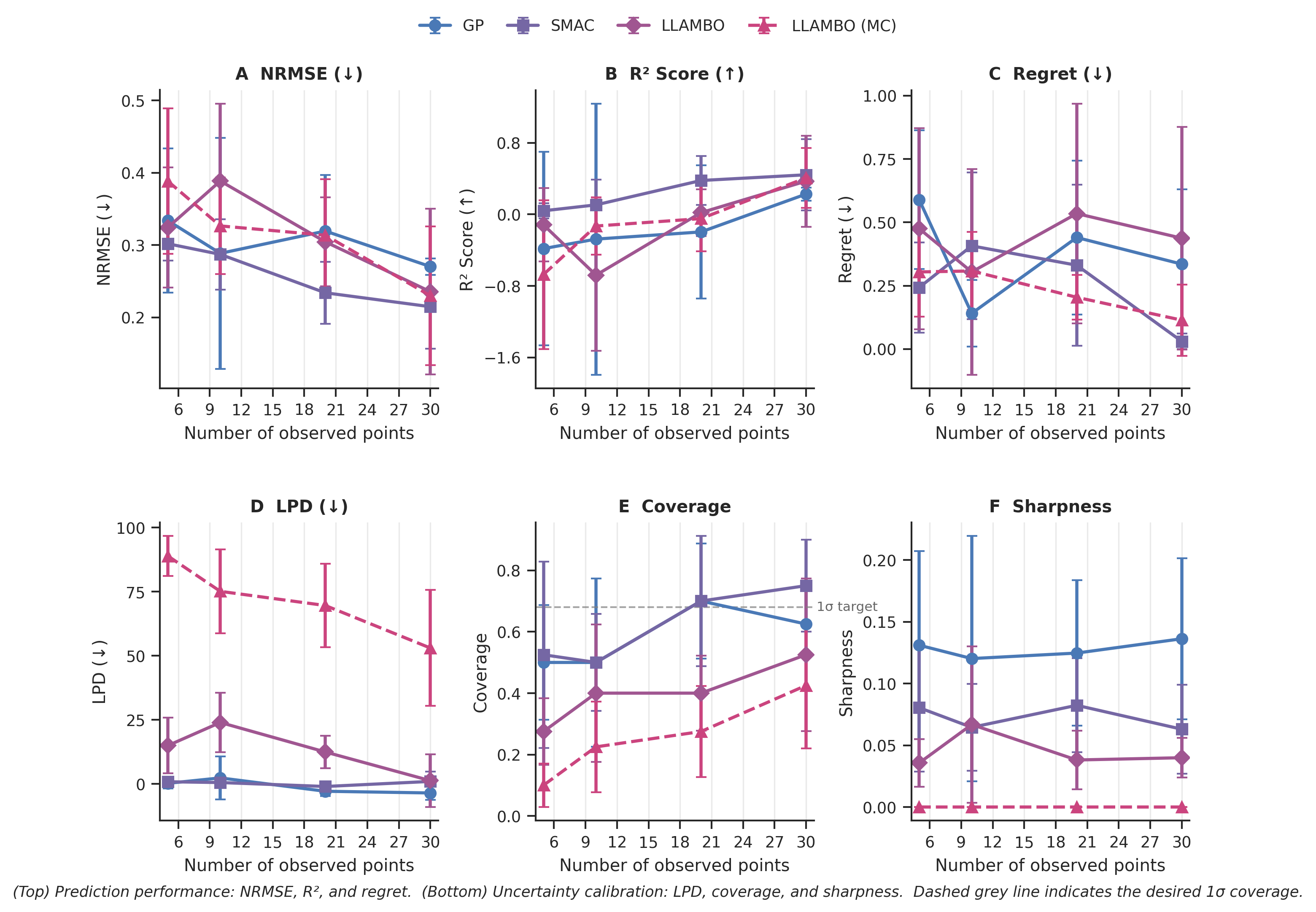}
\caption{\textbf{Discriminative Surrogate models}}
\label{fig:discriminative_surrogate_models}
\end{figure}

Figure~\ref{fig:discriminative_surrogate_models} presents the reproduced evaluation of the discriminative surrogate models 
under varying numbers of observed data points. The metrics follow the original LLAMBO study and assess both predictive accuracy
(NRMSE, $R^2$, regret) and uncertainty quality (LPD, coverage, sharpness). The reproduced results show the same
characteristic pattern: SMAC achieves the strongest pure regression performance with the lowest NRMSE and highest $R^2$,
while Gaussian Processes remain the best calibrated, achieving near-ideal coverage and stable sharpness. 
In contrast, LLAMBO and its Monte Carlo variant exhibit weaker single-task regression performance at low data regimes 
and systematically under-estimated uncertainty, reflected in higher LPD, low coverage, and overly sharp predictive intervals.
However, both LLAMBO variants improve steadily as more observations become available.
These results confirm that LLAMBO is not optimized to serve as a standalone surrogate but instead derives its advantage
from cross-task contextualization and meta-learned priors rather than raw single-task predictive accuracy.

\subsection*{Ablation study}
\begin{figure}[h]
\centering
\includegraphics[width=1\columnwidth]{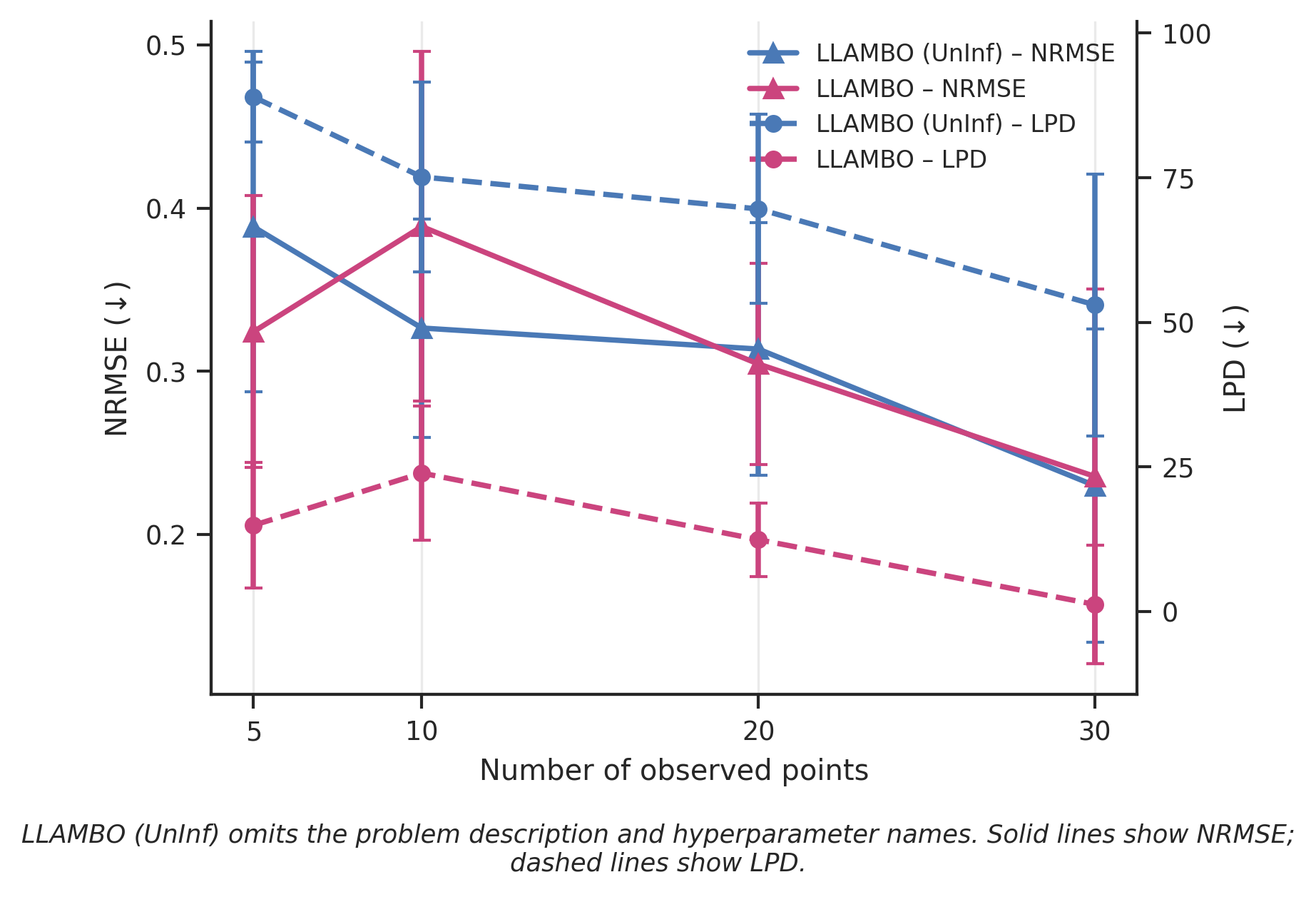}
\caption{\textbf{Ablation study}}
\label{fig:ablation_study}
\end{figure}

Figure~\ref{fig:ablation_study} presents the ablation study comparing the full LLAMBO model with an uninformed variant that excludes
both the problem description and hyperparameter-name embeddings. The reproduced results clearly show that removing textual
context significantly degrades surrogate quality across all data regimes. In terms of predictive accuracy,
the uninformed model exhibits consistently higher NRMSE, with the gap largest when only a small number of observations are available,
indicating impaired generalization and weaker inductive bias. The same trend appears in the uncertainty metrics:
the full LLAMBO model achieves substantially lower LPD, while the uninformed surrogate produces poorly calibrated and overconfident predictions. 
As the number of observations increases, both models improve, but the full LLAMBO surrogate retains a clear advantage. 
This ablation confirms that textual context encoded via the Llama~3.1~70B backbone 
is essential for LLAMBO’s performance, providing semantic structure that enables effective few-shot surrogate modeling.

\subsection*{Candidate point sampling}
\begin{figure}[h]
\centering
\includegraphics[width=1\columnwidth]{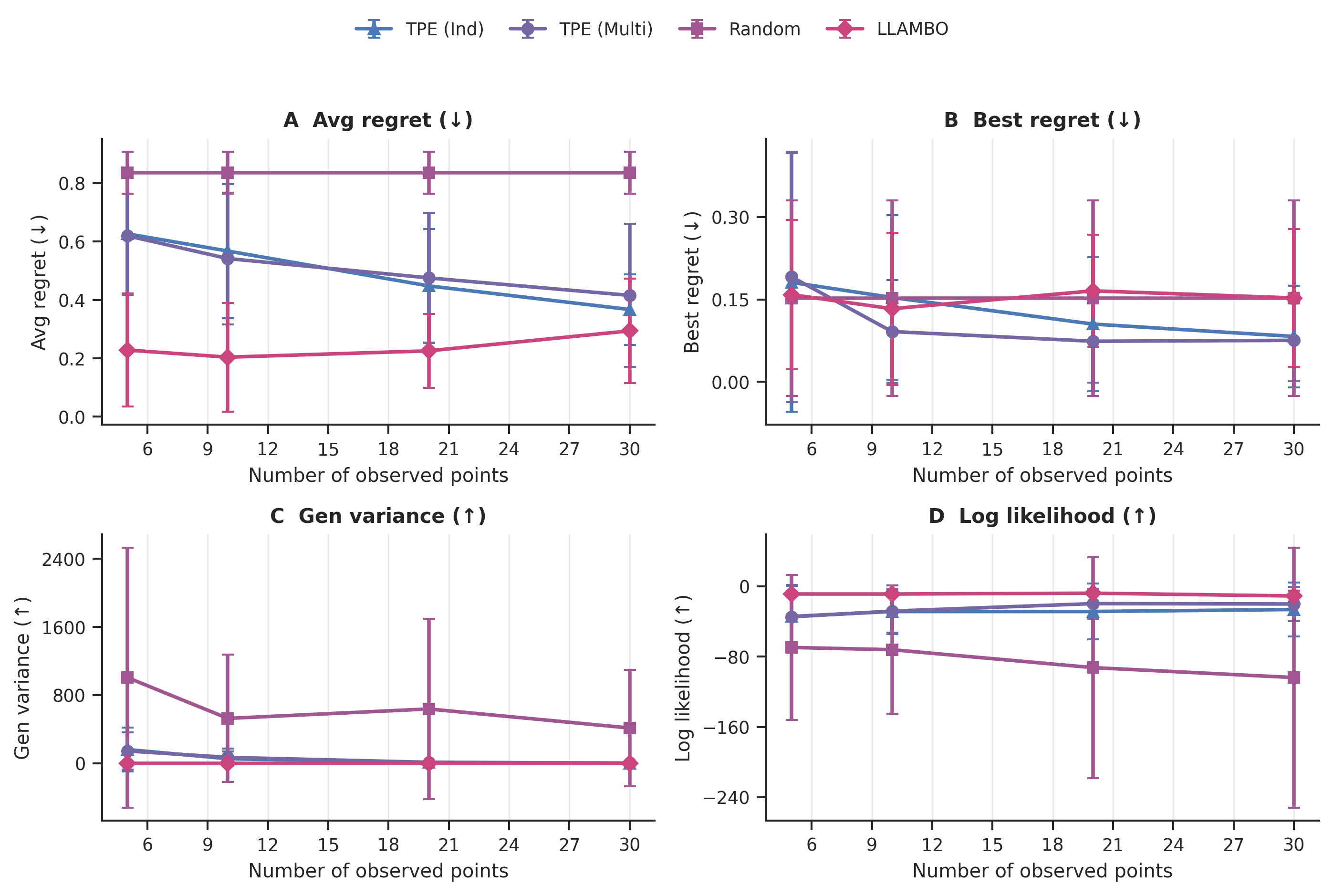}
\caption{\textbf{Candidate Point Sampling}}
\label{fig:candidate_point_sampling}
\end{figure}

Figure~\ref{fig:candidate_point_sampling} shows the reproduced comparison of candidate generation strategies used during acquisition optimization.

We evaluate four methods: Random sampling, TPE with independent marginals, multivariate TPE, and the LLAMBO candidate sampler. 
The reproduced results closely follow the original findings. LLAMBO consistently achieves the lowest average regret across candidate sets,
indicating that it proposes high-quality points aligned with promising regions of the search space. In terms of best-regret performance,
all methods eventually identify strong candidates, but LLAMBO reaches low-regret solutions earlier and with lower variance.
The generalized variance results demonstrate that LLAMBO maintains a balanced level of diversity—avoiding the collapse observed
in TPE-based methods while remaining more targeted than random sampling. Finally, LLAMBO achieves the highest log-likelihood under 
the surrogate density, confirming that its candidates are well-aligned with the learned model structure. Overall, this reproduction 
validates that LLAMBO provides superior candidate generation, combining quality, diversity, and surrogate-consistency more
effectively than existing sampling strategies.

\section*{Discussion}

Our reproduction confirms that the main qualitative claims of LLAMBO hold when GPT-3.5 is replaced by a strong open-weight model. Using Llama~3.1~70B as the backbone, we recover the reported benefits of contextual warmstarting: early-regret is consistently lower, variance across runs is reduced, and the optimization trajectories are more stable than for classical space-filling designs. At the same time, we observe that LLAMBO’s discriminative surrogate is not the strongest single-task regressor—Gaussian Processes and SMAC typically achieve lower NRMSE and better-calibrated uncertainty when trained in isolation. LLAMBO’s advantage instead arises from cross-task semantic priors encoded in the text representations of problem descriptions and hyperparameter names, which improve sample efficiency once contextual information is available.

The ablation study further supports this interpretation. Removing textual context (problem descriptions and hyperparameter semantics) significantly degrades predictive accuracy and calibration, and weakens the warmstarting effect. Nevertheless, the LLAMBO candidate sampler still produces high-quality proposals: across benchmarks, its suggestions are more diverse and better aligned with the underlying objective than those produced by TPE or random sampling, even when the surrogate is not numerically optimal on a single task.

In contrast, our attempts to run the full LLAMBO pipeline with smaller or weaker language models were not successful. Gemma~27B and Llama~3.1~8B, evaluated under the same prompts and parsing logic, frequently returned malformed outputs (invalid JSON, missing hyperparameters) and surrogate scores that did not correlate with observed performance. This led to unstable optimization loops with constraint violations and highly inconsistent rankings of nearly identical candidates. These failures indicate that LLAMBO places non-trivial demands on the underlying LLM: reliable surrogate behaviour requires strong instruction-following, basic numerical reasoning, and reasonably calibrated scoring. In our setup, these properties only emerged robustly with the 70B model, suggesting that—at least with the current prompting scheme—LLAMBO is not yet robust to substantial reductions in model capacity.

\bibliography{references} 

\end{document}